\documentclass[10pt,twocolumn,letterpaper]{article}

\usepackage{iccv}
\usepackage{times}
\usepackage{epsfig}
\usepackage{graphicx}
\usepackage{amsmath}
\usepackage{amssymb}
\usepackage{authblk}
\usepackage{iccv}
\usepackage{times}
\usepackage{epsfig}
\usepackage{graphicx}
\usepackage{amsmath}
\usepackage{amssymb}
\usepackage{amsmath}
\usepackage{amssymb}
\usepackage{subfigure}
\usepackage{algorithm}
\usepackage{algorithmic}
\usepackage{multirow}
\usepackage{times}
\usepackage{epsfig}
\usepackage{graphicx}
\usepackage{amsmath}
\usepackage{amssymb}
\usepackage{subfigure}
\usepackage{algorithm}
\usepackage{algorithmic}
\usepackage{multirow}
\usepackage{booktabs}
\usepackage{xcolor}
\usepackage[accsupp]{axessibility}

\usepackage[breaklinks=true,bookmarks=false]{hyperref}

\iccvfinalcopy 

\def\iccvPaperID{****} 

\ificcvfinal\fi

\newtheorem{lemma}{Lemma}

\begin{document}

\title{Sampling Network Guided Cross-Entropy Method for Unsupervised Point Cloud Registration}

\author[ ]{Haobo Jiang}
\author[ ]{Yaqi Shen}
\author[ ]{Jin Xie$^*$}
\author[ ]{Jun Li}
\author[ ]{Jianjun Qian$^*$}
\author[ ]{Jian Yang}
\affil[ ]{PCA Lab, Nanjing University of Science and Technology, China}
\affil[ ]{\tt\small {\{jiang.hao.bo, syq, csjxie, junli, csjqian, csjyang\}@njust.edu.cn}}

\maketitle

\let\thefootnote\relax\footnotetext{$^*$Corresponding authors}
\let\thefootnote\relax\footnotetext{Haobo Jiang, Yaqi Shen, Jin Xie, Jun Li, Jianjun Qian and Jian Yang are with PCA Lab, Key Lab of Intelligent Perception and Systems for High-Dimensional Information of Ministry of Education, and Jiangsu Key Lab of Image and Video Understanding for Social Security, School of Computer Science and Engineering, Nanjing University of Science and Technology, China.}
\ificcvfinal\thispagestyle{empty}\fi

\begin{abstract}
In this paper, by modeling the point cloud registration task as a Markov decision process, we propose an end-to-end deep model embedded with the cross-entropy method (CEM) for unsupervised 3D registration.
Our model consists of a sampling network  module and a differentiable CEM module.
In our sampling network module, given a pair of point clouds, the sampling network learns a prior sampling distribution over the transformation space. The learned sampling distribution can be used as a ``good" initialization of the differentiable CEM module. In our differentiable CEM module, we first propose a maximum consensus criterion based alignment metric as the reward function for the point cloud registration task. Based on the reward function, for each state, we then construct a fused score function to evaluate the sampled transformations, where we weight the current and future rewards of the transformations. Particularly, the future rewards of the sampled transforms are obtained by performing the iterative closest point (ICP) algorithm  on the transformed state. By selecting the top-k transformations with the highest scores, we iteratively update the sampling distribution. Furthermore, in order to make the CEM differentiable, we use the sparsemax function to replace the hard top-$k$ selection. Finally, we formulate a  Geman-McClure estimator based loss to train our end-to-end registration model.
Extensive experimental results demonstrate the good registration performance of our method on benchmark datasets. 
Code is available at \href{https://github.com/Jiang-HB/CEMNet}{https://github.com/Jiang-HB/CEMNet}.
\end{abstract}

\section{Introduction}
\vspace{-2mm}
Point cloud registration is the problem of finding the optimal rigid transformation (i.e., a rotation matrix and a translation vector) that can align the source point cloud to the target precisely. It plays important roles in a variety of 3D vision applications such as 3D reconstruction~\cite{agarwal2011building,schonberger2016structure,gutim}, Lidar SLAM \cite{deschaud2018imls,zhang2014loam}, 3D object location~\cite{drost2010model,tran2018device,liu2021l3doc}. 
However, various challenges such as outliers and noise interference still hinder its application in the real world.

Owing to discriminative feature extraction of deep neural networks, deep point cloud registration methods \cite{wang2019deep,wang2019prnet,li2020iterative} have shown impressive performance.
Nevertheless, most of their success mainly depends on large amounts of ground-truth transformations for supervised point cloud registration, which greatly increases their training costs. 
To avoid it, recent efforts have been devoted to developing an unsupervised registration model. For example, cycle consistency is used as the self-supervision signal to train the registration models in \cite{groueix2019unsupervised,yang2020mapping}. However, the cycle-consistency loss may not be able to deal with the partially-overlapping case well, since the outliers cannot form a closed loop. In addition, unsupervised point cloud registration methods \cite{wang2020unsupervised,huang2020feature} learn the transformation by minimizing the alignment error (e.g., Chamfer metric) between the transformed source point cloud and the target point cloud.
Nevertheless, for a pair of point clouds with complex geometry, the optimization of the alignment error between them may be difficult and prone to sticking into the local minima.

Inspired by the model based reinforcement learning (RL) \cite{hafner2019learning,hafner2019dream}, in this paper, we propose a novel sampling network guided cross-entropy method for unsupervised point cloud registration. 
By formulating the 3D registration task as a Markov decision process (MDP), it is expected to heuristically search for the optimal transformation by gradually narrowing its interest transformation region through trial and error. 
Our deep unsupervised registration model consists of two modules, i.e., sampling network module and differentiable cross-entropy method (CEM) module. Given a pair of source and target point clouds, the sampling network aims to learn a prior Gaussian distribution over the transformation space, which can provide the subsequent CEM module a proper initialization. With the learned sampling distribution, the differentiable CEM further searches for the optimal transformation by iteratively sampling the transformation candidates, evaluating the candidates, and updating the distribution.  Specifically, in the sampling network module, we utilize the learned matching map with the local geometric feature for the mean estimation of the sampling distribution and exploit the global feature for variance estimation, respectively. 
In the CEM module,  a novel fused score function combining the current and the future rewards is designed to evaluate each sampled transformation. Particularly,  the future reward is estimated by performing the ICP algorithm on the transformed source point cloud and the target point cloud. 
In addition, the differentiability in our CEM module is achieved by softening the sorting based top-$k$ selection with a differentiable sparsemax function. 
Finally, we formulate a scaled Geman-McClure estimator \cite{zhou2016fast} based loss function to train our model, where the sublinear convergence speed for the outliers can weaken the negative impact on registration precision from the outliers. 

To summarize, our main contributions are as follows:
\vspace{-2mm}
\begin{itemize}
	\item We propose a novel end-to-end cross-entropy method based deep model for unsupervised point cloud registration, where a prior sampling distribution is predicted by a sampling network to quickly focus on the promising searching region.
\vspace{-2mm}
	\item In the cross-entropy method, we design a novel ICP driven fused reward function for accurate transformation candidate evaluation and propose a spasemax function based soft top-$k$ selection mechanism for the differentiability of our model.
	
\vspace{-2mm}
	\item Compared to the unsupervised or even some fully supervised deep methods, our method can obtain outstanding performance on extensive benchmarks. 
\end{itemize}

\section{Related Work}
\noindent\textbf{Traditional point cloud registration algorithms.}  
In the traditional 3D registration field, a lot of research has focused on the Iterative Closest Point algorithm (ICP) \cite{besl1992method} and its variants. The original ICP repeatedly performs correspondence estimation and transformation optimization to find the optimal transformation. However, ICP with improper initialization easily sticks into the local minima dilemma and the sensitivity to the outliers also degrades its performance in the partially overlapping case.
To avoid it, Go-ICP \cite{yang2013go} globally searches for the optimal transformation via integrating the branch-and-bound scheme into the ICP. 
Furthermore, Chetverikov \textit{et al.} \cite{chetverikov2002trimmed} proposed a trimmed ICP (TrICP) algorithm to handle the partial case, where the least square optimization is only performed on partial minimum square errors rather than the all. 
Moreover, other variants such as \cite{sharp2002icp,fitzgibbon2003robust,bae2008method, gressin2013towards, deng2018ppfnet} also present competitive registration perfromance.
In addition, RANSAC based methods \cite{fischler1981random,chum2003locally, chum2005matching,torr2000mlesac,tordoff2005guided,dong2020registration} also have been extensively studies. 
Among them, one representative method is the 4PCS \cite{aiger20084, theiler2014fast, theiler2014keypoint} which determines the correspondence from the sampled four-point sets that are nearly co-planar via comparing their intersectional diagonal ratios. 
Furthermore, Super4PCS \cite{mellado2014super}, an improved 4PCS method, is proposed to largely reduce the computational complexity of the congruent four-point sets sampling. Compared with the quadratic time complexity of points in 4PCS, Super4PCS only requires linear complexity, which greatly promotes its application in the real world. 
In addition, other RANSAC variants, including \cite{mohamad2014generalized, mohamad2015super, xu2019pairwise, huang2017v4pcs, ge2017automatic, le2019sdrsac} also show good alignment results.

\vspace{-0mm}
\noindent\textbf{Learning-based point cloud registration algorithms.} 
In recent years, deep learning based point cloud registration methods have received widespread attention.
PPFNet method \cite{deng2018ppfnet} is proposed to utilize the PointNet \cite{qi2017pointnet} to extract the feature of the point patches and then perform the RANSAC for finding corresponding patches.
DCP \cite{wang2019deep} calculates the rigid transformation via the singular value decomposition (SVD) where the correspondence is constructed through learning a soft matching map.
RPM-Net \cite{yew2020rpm} also realizes the registration based on the matching map generated from the Sinkhorn layer and annealing.
\cite{yuan2020deepgmr} aligns the point cloud pair by minimizing the KL-Divergence between two learned Gaussian Mixture Models. 
In addition, ReAgent \cite{bauer2021reagent} is proposed to combine the imitation learning and model-free reinforcement learning (i.e., proximal policy optimization \cite{schulman2017proximal}) for registration agent training. Instead,  our method utilizes the model-based cross-entropy method for registration with the constructed MDP model.
\cite{jiang2021planning} proposes to search for the optimal solution by planning with a learned latent dynamic model. Although it's unsupervised and has a faster inference speed, the approximation error in its latent reward and translation networks may potentially degrade its registration precision.
Moreover, \cite{groueix2019unsupervised,yang2020mapping} propose to apply the cycle consistency as the supervision signal into the point cloud registration for the unsupervised learning. However, since the non-overlapping region cannot construct a closed-loop, they also may not handle the partial case well.
In addition, other learning-based methods, including \cite{wang2019prnet,li2020iterative, choy2020deep, pais20203dregnet,li2020unsupervised,li2019pc,zhu2020reference} also present impressive performance.

\begin{figure*}[t]
	\centering  
	\subfigure{
		\includegraphics[width=1.0\textwidth]{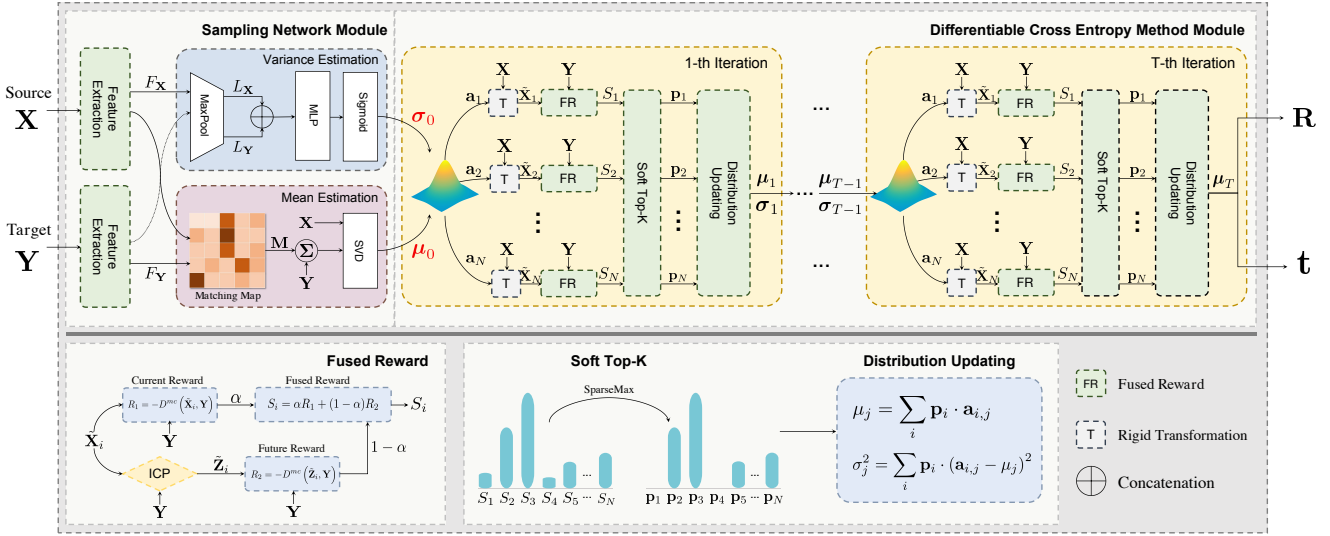}}
	\caption{
		The pipeline of the proposed sampling network guided cross-entropy method (CEM) for point cloud registration. Our framework mainly contains two cascaded modules: (a) Sampling network module: Given the source and target point clouds $\left\{\mathbf{X},\mathbf{Y}\right\}$, we extract their per-point features $F_\mathbf{X}$ and the global feature $L_\mathbf{Y}$ which are used to predict the mean $\boldsymbol{\mu}_0$ and the standard deviation $\boldsymbol{\sigma}_0$ of the initial sampling distribution ($\mathcal{N}\left(\boldsymbol{\mu}_0,\boldsymbol{\sigma}\right)$) for subsequent CEM block. 
		(b) Differentiable CEM module: In the $t$-th iteration ($1\leq t\leq T$), it alternatively performs the
		sampling transformation candidates $\left\{\mathbf{a}_1,..., \mathbf{a}_N\right\}$; evaluating sampled transformations (obtaining scores $\left\{{S}_1, ..., {S}_N\right\}$) through our designed fused score function which combines the current and future rewards; mapping the obtained scores to a sparse distribution $\left\{\mathbf{p}_1,...,\mathbf{p}_N\right\}$ with the differentiable sparsemax function; updating the sampling distribution via Eq.~\ref{update2}. In the last iteration, we use the expected value of the sampling distribution to estimate the optimal transformation.}
	\label{framework}
	\vspace{-3mm}
\end{figure*}

\section{Approach}
\subsection{Problem Setting}
For the point cloud registration task, given the source point cloud $\mathbf{X}=\{\mathbf{x}_i\in\mathbb{R}^3\mid i=1,...,N\}$ and the target point cloud $\mathbf{Y}=\{\mathbf{y}_j\in\mathbb{R}^3 \mid j=1,...,M\}$, we aim to recover the rigid transformation containing a rotation matrix $\mathbf{R} \in SO\left(3\right)$ and a translation vector $\mathbf{t}\in\mathbb{R}^3$ for aligning the two point clouds. 
In this work, we formulate 3D point cloud registration as a Markov decision process (MDP), which contains a state set, an action set, a state transition function, and a reward function. 


\noindent\textbf{State space}. We define the set of point cloud pairs as the state space $\mathcal{S}$, that is $\mathbf{s}=\left\{{\mathbf{X}}, {\mathbf{Y}}\right\}\in\mathcal{S}$.

\noindent\textbf{Action space}. We denote the rigid transformation space $SE(3)$ ($SO(3) \times \mathbb{R}^{3}$) as the action space $\mathcal{A}$. 
To reduce the dimension of the search space, we utilize the Euler angle representation $\boldsymbol{e}=[e_1,e_2,e_3] \in [-\pi, \pi]^3$ to encode the rotation $\mathbf{R} \in SO(3)$. 
Thus, the action $\mathbf{a} \in \mathcal{A}$ is represented by $\mathbf{a}=[\mathbf{e}, \mathbf{t}]$ and $\mathbf{R}(\mathbf{e})$ is denoted as the corresponding rotation matrix of the Euler angle $\mathbf{e}$.

\noindent\textbf{State transition function}. After executing the action $\mathbf{a}=\left[\mathbf{e}, \mathbf{t}\right]$ at the current state $\mathbf{s}=\left\{{\mathbf{X}}, {\mathbf{Y}}\right\}$, we predict the next registration state $\mathbf{s}^\prime=\left\{{\tilde{\mathbf{X}}}, {\mathbf{Y}}\right\}$ with the state transition function $\mathcal{T}$, that is:
\begin{equation}
\mathbf{s}^\prime=\mathcal{T}\left(\mathbf{s}, \mathbf{a}\right)=\left\{ \left\{\mathbf{R}\left(\mathbf{e}\right)\mathbf{x}_i + \mathbf{t}\right\}_{i=1}^N, \mathbf{Y}\right\}=\left\{\tilde{\mathbf{X}}, {\mathbf{Y}}\right\}.
\end{equation}



\noindent\textbf{Reward function}. We evaluate the effect of executing the action $\mathbf{a}$ at the state $\mathbf{s}$ with the reward function $R\left(\mathbf{s}, \mathbf{a}\right)$  in the point cloud registration task.  
 In order to handle the partially overlapping case, we define a maximum
consensus criterion based alignment metric $D^{mc}$ as below to quantify the reward, that is, $R\left(\mathbf{s}, \mathbf{a}\right)=-D^{mc}\left(\mathcal{T}\left(\mathbf{s},\mathbf{a}\right)\right)=-D^{mc}\left(\tilde{\mathbf{X}}, \mathbf{Y}\right)$. The lower the alignment error between ${\tilde{\mathbf{X}}}$ and ${\mathbf{Y}}$, the higher the reward obtained by that action:
\begin{equation}\label{reward}
		\begin{split}
			D^{mc}\left(\tilde{\mathbf{X}}, \mathbf{Y}\right)=2 -
			&\frac{1}{N}\sum_{\tilde{\mathbf{x}}_i\in{\tilde{\mathbf{X}}}}{\rho_{\varepsilon}\left(d_{\tilde{\mathbf{x}}_i,\mathbf{Y}}\right)\left(1-\frac{d_{\tilde{\mathbf{x}}_i,\mathbf{Y}}}{\varepsilon}\right)} -\\ & \frac{1}{M}\sum_{\mathbf{y}_j\in{\mathbf{Y}}}{\rho_{\varepsilon}\left(d_{\mathbf{y}_j,\tilde{\mathbf{X}}}\right)\left(1-\frac{d_{\mathbf{y}_j,\tilde{\mathbf{X}}}}{\varepsilon}\right)},
	\end{split}
\end{equation} where $d_{\tilde{\mathbf{x}}_i,\mathbf{Y}}=\min_{\mathbf{y}_j}\left\|\tilde{\mathbf{x}}_i-\mathbf{y}_j\right\|_2$ denotes the shorest distance between the point $\tilde{\mathbf{x}}_i$ and the point cloud $\mathbf{Y}$ and $\rho_{\varepsilon}\left(x\right)=1\left\{x \leq \varepsilon\right\}$ is the indicator function, where the threshold  $\varepsilon > 0$ is a hyper-parameter that specifies the maximum allowable distance for inlier. $D^{mc}$ measures the alignment error via counting the number of overlapping point pairs, which can effectively relieve the effect of outliers. 
Furthermore, each count in $D^{mc}$ is weighted with	$1-d_{\tilde{\mathbf{x}}_i,\mathbf{Y}}/{\varepsilon}$. Compared to the discrete count \cite{le2019sdrsac}, such weighting operation enforces the overlapping pair to be as close as possible rather than just no more than the maximum  allowable distance. 
\begin{lemma}
	If the threshold $\varepsilon$ is no less than the maximum of distances between all closest point pairs, that is, $\max\left\{\max_{i} d_{\tilde{\mathbf{x}}_i,\mathbf{Y}}, \max_jd_{{\mathbf{y}}_j,\tilde{\mathbf{X}}}\right\} \leq \varepsilon<\infty$,
	 $D^{mc}$ is proportional to the Chamfer metric $D^{cd}$ with ratio $1/\varepsilon$, i.e., $D^{mc}\left(\tilde{\mathbf{X}},\mathbf{Y}\right)=1/\varepsilon\cdot D^{cd}\left(\tilde{\mathbf{X}},\mathbf{Y}\right)$.
\end{lemma} 
The proof is provided in Appendix A.

Based on the constructed MDP, given a source and target point cloud pair $\mathbf{s}=\left\{\mathbf{X},\mathbf{Y}\right\}$, the point cloud registration task aims to find the optimal action $\mathbf{a}^*$  by maximizing its reward as below:
\begin{equation}\label{target}
	\begin{split}
		\mathbf{a}^* = \underset{\mathbf{a}\in\mathcal{A}}{\operatorname{argmax}}R\left(\mathbf{s},\mathbf{a}\right) 
		= \underset{\mathbf{a}\in\mathcal{A}}{\operatorname{argmin}}D^{mc}\left(\mathcal{T}\left(\mathbf{s},\mathbf{a}\right)\right). 
	\end{split}
\end{equation}

\subsection{Differentiable Cross-Entropy Method based Deep Registration Network}
To solve the optimization problem in Eq.~\ref{target}, we propose a novel CEM based unsupervised cloud registration framework.
As shown in Fig.~\ref{framework}, our deep model contains two cascaded modules, including a sampling network module and a differentiable CEM module. 
With the source and target point cloud pair as the input, the sampling network module  predicts a sampling distribution over the transformation space (Section~\ref{sec:sample}). Guided by this sampling distribution, the CEM module searches the optimal transformation by iteratively narrowing the search space through the transformation sampling, transformation evaluation, and distribution updating. 
The optimal transformation can be estimated with the mean of the elite transformations in the last iteration (Section~\ref{sec:cem}).

\subsubsection{Sampling Network Module}
\label{sec:sample}
Given the source and target point cloud pair $\mathbf{s}=\left\{\mathbf{X},\mathbf{Y}\right\}$, the traditional CEM randomly samples candidates from a pre-defined sampling distribution. Instead,  we aim to sample them with a learned Gaussian distribution $\left\{\boldsymbol{\mu}_0,\boldsymbol{\sigma}_0\right\}=\mathcal{P}\left(\mathbf{s};\mathbf{w}\right)$ parameterized by the deep neural network with parameters $\mathbf{w}$, 
so that  given a registration state, it can increase the chance that sampled candidates fall into the neighborhood of the optimal transformation. As shown in Fig.~\ref{framework},  we utilize the Dynamic Graph CNN (DGCNN) \cite{wang2019dynamic} to extract the local geometric features $F_\mathbf{X}\in\mathbb{R}^{N\times P}$ and $F_\mathbf{Y}\in\mathbb{R}^{M\times P}$ via constructing the $k$-NN graph for each point in $\left\{\mathbf{X},\mathbf{Y}\right\}$. With the learned point embeddings, we further generate a matching map $\mathbf{M}\in\mathbb{R}^{N\times M}$ with the softmax function:
\begin{equation}
	\begin{split}
		\mathbf{M}_{i,j} = \operatorname{softmax}\left(F_{\mathbf{Y}} F_{\mathbf{x}_i}^\top\right)_j,
	\end{split}
\end{equation} where $F_{\mathbf{x}_i}\in F_{\mathbf{X}}$ denotes the feature of the point $\mathbf{x}_i$ and the  element $\mathbf{M}_{i,j}$ denotes the matching probability of the points $\mathbf{x}_i$ and $\mathbf{y}_j$. 
With the map $\mathbf{M}$, we can estimate the ``matching'' point $\hat{\mathbf{y}}_i$ of $\mathbf{x}_i$: 
\begin{equation}
	\begin{split}
\hat{\mathbf{y}}_i = \sum_{j=1}^M \mathbf{M}_{i,j} \cdot \mathbf{y}_j \in \mathbb{R}^3.
	\end{split}
\end{equation}  
Once the correspondences  $\left\{\left(\mathbf{x}_i,\hat{\mathbf{y}}_i\right)\mid i=1,\ldots,N\right\}$ are obtained, we employ the singular value decomposition (SVD) to calculate the transformation $\left\{\mathbf{R},\mathbf{t}\right\}$. The expected value  
 $\boldsymbol{\mu}_0$ of the Gaussian distribution is equal to the concatenated vector $\left[\mathbf{e},\mathbf{t}\right] \in \mathbb{R}^6$, where $\mathbf{e}$ is the corresponding Euler angle of $\mathbf{R}$. For the prediction of the standard deviation $\boldsymbol{\sigma}_0$, we first extract the global features of the point cloud pair $L_{\mathbf{X}}\in\mathbb{R}^P$ and $L_{\mathbf{Y}}\in\mathbb{R}^P$ via the maxpooling operation, that is, $L_{\mathbf{X}}=\operatorname{MaxPool}\left(F_{\mathbf{X}}\right)$. Then, we feed the concatenated global feature $\left[L_{\mathbf{X}}, L_{\mathbf{Y}}\right]$ into the 3-layer MLP followed by a sigmoid function to output $\boldsymbol{\sigma}_0$.

\begin{algorithm}
	\caption{CEM based deep registration framework}
	\label{alg:algorithm}
	\textbf{Input:} state $\mathbf{s}=\left\{\mathbf{X},\mathbf{Y}\right\}$, sampling network $\mathcal{P}\left(\mathbf{s};\mathbf{w}\right)$, iteration times $T$, action candidate numbers $N$,  iteration times $M$ of using future reward. 
	
	\begin{algorithmic}[1] 
		\STATE Initilize sampling distribution $\left\{\boldsymbol{\mu}_0, \boldsymbol{\sigma}_0\right\}=p_0\left(\mathbf{s};\mathbf{w}\right).$
		\FOR{$t = 0:T-1$}
		\FOR{$i = 1:N$}
		\STATE Sample transformation $\mathbf{a}_{t}^i\sim  \mathcal{N}\left(\boldsymbol{\mu}_t, \boldsymbol{\sigma}_t^2\mathbb{I}\right)$.
		\STATE Perform $\mathbf{a}_{t}^i$ at the state $\mathbf{s}$ and obtain the next state $\mathbf{s}'_{i}$ and reward $R\left(\mathbf{s}, \mathbf{a}_{t}^i\right)$.
		\IF{$t\geq M$}
		\STATE \textcolor{gray}{\% Only-current reward.}
		\STATE Score $S\left(\mathbf{s}, \mathbf{a}_{t}^i\right)=R\left(\mathbf{s}, \mathbf{a}_{t}^i\right)$.
		\ELSE
		\STATE \textcolor{gray}{\% Fused reward.}
		\STATE Perform ICP  to registrate $\mathbf{s}'_{i}$ and obtain the predicted transformation $\mathbf{a}_{icp}^i$.
		\STATE Calculate the $R(\mathbf{s}_i', \mathbf{a}_{icp}^i)$ as the future reward.
		\STATE Calculate fused score $S\left(\mathbf{s}, \mathbf{a}_{t}^i\right)$ via Eq.~\ref{score}.
		\ENDIF
		\ENDFOR
		\STATE Calculate the sparsemax based weight vector via Eq.~\ref{sparse} and update the distribution via Eq.~\ref{update2}.
		\ENDFOR	
	\end{algorithmic}
\textbf{Return:} $\boldsymbol{\mu}_T$
	\label{alg1}
\end{algorithm}

\vspace{-5mm}
\subsubsection{Differentiable Cross-entropy Method Module}
\label{sec:cem}
Guided by the prior sampling distribution learned by the network guidance module, the CEM model further iteratively searches for the optimal transformation by extensive trial and error. 
During the iterative process, we propose a novel fused score function combing the current reward and the ICP based future reward for accurate transformation evaluation.
Furthermore, we also propose a novel differentiable CEM for the end-to-end training, where the hard top-$k$ operation is replaced by a differentiable sparsemax function. 
Specifically, in the $t$-th iteration ($0\leq t < T$), it executes the three steps as below:

\noindent\textbf{(a) Transformation candidates sampling.} In this step, we randomly sample $N$ transformation candidates $\mathcal{C}_t=\left\{\mathbf{a}_t^{i}\mid i =1,\ldots,N\right\}$ from a Gaussian distribution $g_t = \mathcal{N}\left(\boldsymbol{\mu}_t, \boldsymbol{\sigma}^2_t\mathbb{I}\right)$.
Note that we exploit the prior distribution $\left\{\boldsymbol{\mu}_0,  \boldsymbol{\sigma}_0\right\}=\mathcal{P}\left(\mathbf{s};\mathbf{w}\right)$ predicted by the sampling network for the first iteration so that it can quickly focus on the most promising transformation region for the optimal solution searching. For differentiability, we utilize the reparameterization trick to sample each candidate, that is, $\mathbf{a}_t^i = \boldsymbol{\mu}_t + \epsilon_i\cdot\boldsymbol{\sigma}_t$, where $\epsilon_i\sim\mathcal{N}\left(\boldsymbol{0},\boldsymbol{1}\mathbb{I}\right)$. 

\begin{figure}[t]
	\centering  
	\subfigure{
		\includegraphics[width=1\columnwidth]{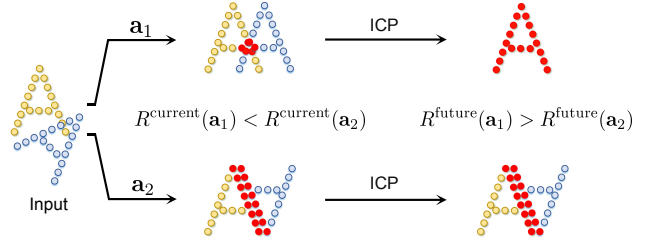}}
	\caption{After performing the actions $\mathbf{a}_1$ and $\mathbf{a}_2$ on the source point cloud (blue), the transformed source point cloud via $\mathbf{a}_2$ has more overlapping pairs (red) with the target point cloud (yellow) than those via $\mathbf{a}_1$ (i.e., larger current reward, $R^{\text{current}}(\mathbf{a}_1)<R^{\text{current}}(\mathbf{a}_2)$). However, the action $\mathbf{a}_1$ can lead to a better pose for future alignment than $\mathbf{a}_2$ and obtain larger overlappings after performing the ICP alignment. Thus, the future reward of the action $\mathbf{a}_1$ is higher than that of the action $\mathbf{a}_2$ ($R^{\text{future}}(\mathbf{a}_1)>R^{\text{future}}(\mathbf{a}_2)$).
	}
	\label{fig:rethink2}
	\vspace{-3mm}
\end{figure}

\noindent\textbf{(b) Fused reward based transformation evaluation.} After sampling the transformation candidates, based on our constructed MDP, we can obtain the corresponding rewards $\left\{R\left(\mathbf{s},\mathbf{a}_t^i\right)\mid 1\leq i \leq N\right\}$ for the sampled transformation candidates. 
In the traditional CEM, the reward is directly used to score each transformation, i.e., the score function $S\left(\mathbf{s},\mathbf{a}_t^i\right)=R\left(\mathbf{s},\mathbf{a}_t^i\right)$.
However, a good transformation candidate needs to own not only the high current reward but also the high future reward.
For example, as shown in Fig.~\ref{fig:rethink2}, although the transformation $\mathbf{a}_2$ can temporarily obtain a higher reward than $\mathbf{a}_1$ due to the more overlapping pairs, the resulting registration state by performing $\mathbf{a}_2$ has a worse pose than the state by performing $\mathbf{a}_1$ intuitively. Thus, for a transformation, just focusing on the current reward and ignoring the future reward of a transformation may mislead the evolution of the CEM and cause it to stick into the local optima. 
Therefore, for each transformation, we need to define a  future reward  to determine whether the transformd source and target point clouds have good poses at the next registration state.

In our method, we utilize the ICP algorithm to quantify the future reward of a transformation.
Specifically,  we first perform each sampled transformation at the state $\mathbf{s}$ and obtain a set of states $\left\{\mathbf{s}_1',\ldots,\mathbf{s}_N'\right\}$ where $\mathbf{s}_i'=\mathcal{T}\left(\mathbf{s},\mathbf{a}_t^i\right)=\left\{\tilde{\mathbf{X}}_i,\mathbf{Y}\right\}$.
Then, we use ICP to predict the transformation $\mathbf{a}_{{icp}}^i$ for the registration state $\mathbf{s}_i'$, and obtain the corresponding next state $\left\{\tilde{\mathbf{Z}}_{i}, \mathbf{Y}\right\}$ and the reward $R\left(\mathbf{s}_i',\mathbf{a}_{icp}^i\right)=-D^{mc}\left(\tilde{\mathbf{Z}}_{i}, \mathbf{Y}\right)$. 
Finally,  we use this reward as the future reward. The higher reward  means $\mathbf{a}_t^i$ may lead to a better pose at the next registration state.
After that, we utilize the weighted sum of the current and future rewards to score each action:
\begin{equation}
	\begin{split}\label{score}
		S(\mathbf{s},\mathbf{a}_t^i)=\alpha\cdot R(\mathbf{s},\mathbf{a}_t^i) + (1-\alpha)\cdot R(\mathbf{s}_i', \mathbf{a}_{icp}^i),
	\end{split}
\end{equation}
where $\alpha\in[0,1]$ is the hyper-parameter for balancing the weights of the current and future rewards. Note that benefitting from the CUDA programming, we accelerate the inference speed with the batch ICP consisting of the batch SVD and point-wise parallelization for closest point search.

\noindent\textbf{(c) Sparsemax based distribution updating.} 
In this step, we aim to use the elite transformations to update the sampling distribution so that more ``better" candidates are expected to be sampled in the next iteration.
In the traditional CEM, the top-$k$ transformations $\mathcal{I}_t$ with the highest scores are used as the elites to guide the distribution updating $g_{t+1}=\mathcal{N}\left(\boldsymbol{\mu}_{t+1},\boldsymbol{\sigma}_{t+1}^2\mathbb{I}\right)$ by the updating formula:
\begin{equation}\label{update1}
	\begin{split}
		{\mu}_{t+1,i}=\frac{1}{k}\sum_{\mathbf{a}\in\mathcal{I}_t}\mathbf{a}_i,\ \boldsymbol{\sigma}_{t+1,i}^2=\frac{1}{k}\sum_{\mathbf{a}\in\mathcal{I}_t}\left(\mathbf{a}_i-\mu_{t+1,i}\right)^2,
	\end{split}
\end{equation}where $i=1,...,6$. In fact, Eq.~\ref{update1} is the closed-form solution of the maximum likelihood estimation problem over the elites, that is, $\left\{\boldsymbol{\mu}_{t+1},\boldsymbol{\sigma}_{t+1}\right\}={\operatorname{argmax}_{\boldsymbol{\mu},\boldsymbol{\sigma}}}\sum_{\mathbf{a}\in\mathcal{I}_t}f\left(\mathbf{a};\boldsymbol{\mu},\boldsymbol{\sigma}\right)$, where $f(\cdot;\boldsymbol{\mu},\boldsymbol{\sigma})$ denotes the probability density function of $\mathcal{N}\left(\boldsymbol{\mu},\boldsymbol{\sigma}^2\right)$.
By re-fitting the sampling distribution with the top-$k$ samples, the updated sampling distribution tends to focus on the more promising transformation region.
However, since the sorting in the top-$k$ operation is non-differentiable, such distribution updating method cannot be directly used for end-to-end training directly. Therefore, we propose to soften the hard top-$k$ selection via a differentiable sparsemax function \cite{martins2016softmax}, which is different from LML layer used in \cite{amos2020differentiable}.
For a clear insight, we first rewrite  Eq.~\ref{update1} as:
\begin{equation}\label{update2}
		\small{\begin{split}
			{\mu}_{t+1,i}=\sum_{\mathbf{a}\in\mathcal{C}_t}p\left(\mathbf{a}\right)  \mathbf{a}_i,\ {\sigma}_{t+1,i}^2=\sum_{\mathbf{a}\in\mathcal{C}_t}p\left(\mathbf{a}\right)  \left(\mathbf{a}_i-\mu_{t+1,i}\right)^2
	\end{split}}
\end{equation}
where $p\left(\mathbf{a}\right)=\frac{1}{k}\cdot 1\left\{\mathbf{a} \in \mathcal{I}_t \right\}$ is the weight assigned to each candidate $\mathbf{a}_t^i \in \mathcal{C}_t$ for distribution updating and the sum of all elements in $\mathbf{p}=\left[p\left(\mathbf{a}_t^1\right),..., p\left(\mathbf{a}_t^N\right)\right]$ is 1. 
Next, given the score vector $\mathbf{q}=\left[S\left(\mathbf{s},\mathbf{a}_t^1\right),...,S\left(\mathbf{s},\mathbf{a}_t^N\right)\right]$ of all candidates, sparsemax based weight vector $\tilde{\mathbf{p}}$ is defined as the solution of the following minimization problem:
\begin{equation}\label{sparse1}
		\begin{split}
			\tilde{\mathbf{p}}={\rm sparsemax}\left(\mathbf{q}\right)=\underset{\mathbf{z} \in \Delta^{N-1}}{\operatorname{argmin}}\|\mathbf{z}-\mathbf{q}\|^{2}
	\end{split},
\end{equation}
where $\Delta^{N-1}=\left\{\mathbf{z}\in\mathbb{R}^N\mid \sum_i\mathbf{z}_i=1, \mathbf{z}_i\geq0 \right\}$ is a ($N$-1)-dimensional simplex. Eq.~\ref{sparse1} has a closed-form solution, that is ${\rm sparsemax}_i\left(\mathbf{q}\right)=\left[S\left(\mathbf{s},\mathbf{a}_t^i\right)-\tau\left(\mathbf{q}\right)\right]_+$, and the threshold function $\tau\left(\mathbf{q}\right)$ can be expressed as:
\begin{equation}\label{sparse}
	\begin{split}
		\tau\left(\mathbf{q}\right) = \frac{\sum_{j\leq \mathcal{M}\left(\mathbf{q}\right)}\mathbf{q}_{(j)}-1}{\mathcal{M}\left(\mathbf{q}\right)}
	\end{split},
\end{equation}
where $\mathbf{q}_{\left(j\right)}$ denotes the $j$-largest value in the vector $\mathbf{q}$ and $\mathcal{M}\left(\mathbf{q}\right)=\operatorname{max}\left\{m \mid 1 + m\mathbf{q}_{\left(m\right)} > \sum_{j\leq m }\mathbf{q}_{\left(j\right)} \right\}$ \cite{martins2016softmax}. The more details about the function $\tau$ and the Jacobian of sparsemax for back propagation can be seen in Appendix B.
Note that the sparsemax function maps the scores of candidates to a probability distribution whose some elements will be zero if the scores of the corresponding candidates are too low. 
Compared to the softmax function, the sparsemax function can adaptively ignore the negative effect of the bad candidates, which better simulates the hard top-$k$ operation. 
Moreover, instead of assigning the same weight ($1/k$) to the candidates, the sparsemax function assigns the higher weights to the better candidates. Thus, the updated distribution can focus on the actions with the higher scores and more potential candidates in the next iteration can be sampled. 
After $T$ iterations, we use the expected value of the sampling distribution to estimate the optimal transformation. The CEM based registration algorithm is outlined in Algorithm \ref{alg:algorithm}.

\begin{figure*}[t]
	\centering 
	\subfigure{
		\includegraphics[width=1.0\textwidth]{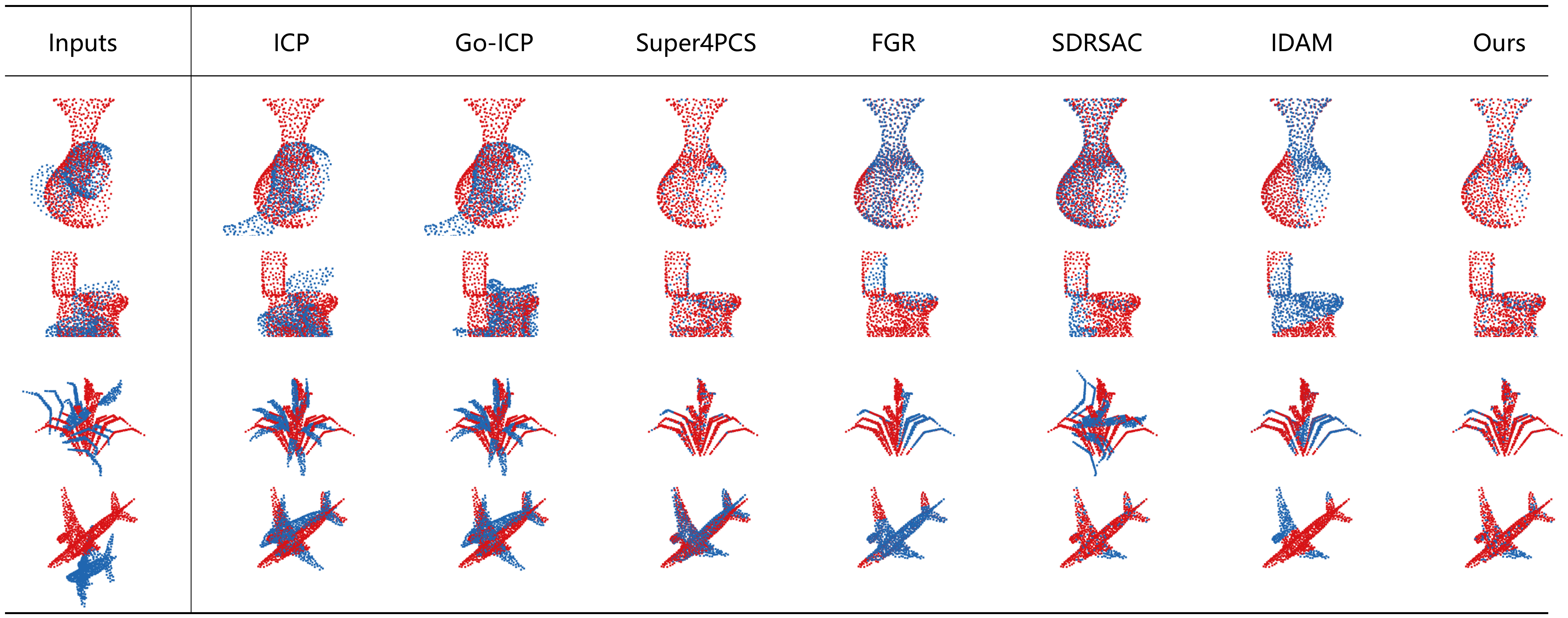}}
	\caption{
		Qualitative registration examples on partially overlapping \textit{ModelNet40} dataset.
	}
	\label{fig:vis}
\end{figure*}

\subsection{Loss Function}
\vspace{-1mm}
\label{sec:loss}
Under the unsupervised setting, we utilize the alignment error between the transformed source point cloud and target point cloud rather than the ground-truth transformation for model training. 
To handle the partially-overlapping case well, a proper robust loss function insensitive to the outliers is desired. 
In this paper, we integrate the scaled Geman-McClure estimator $\rho_\mu$ \cite{zhou2016fast} into our loss function:
\begin{equation}
	\label{smooth}
	\begin{split}
		\rho_{\mu}\left(x\right) = \frac{\mu\cdot x^2}{\mu + x^2},
	\end{split}
\end{equation}
where the scaled Geman-McClure estimator puts the least-squared penalty on the inliers while assigns the sublinear penalty on the outliers \cite{zhou2016fast}. 
Thus,  the  resulting loss function can effectively weaken the alignment error from the outliers.
The hyper-parameter $\mu$ determines the range of the inliers.
Given a training dataset $\mathcal{D}=\left\{\left(\mathbf{X}_i, \mathbf{Y}_i\right)\right\}$, the loss function $\mathcal{L}\left(\mathbf{w}\right)$ is defined as below:
\begin{equation}
	\label{smooth}
	\begin{aligned}
		\mathcal{L}\left(\mathbf{w}\right)=\mathbb{E}_{\mathbf{X},\mathbf{Y}\sim\mathcal{D}}\bigg[
		&\frac{1}{|\mathbf{X}|}
		\sum_{\tilde{\mathbf{x}}_i\in{\tilde{\mathbf{X}}\left(\mathbf{w}\right)}}{\rho}_{\mu}\left(d_{\tilde{\mathbf{x}}_i,\mathbf{Y}}\right) +\\ &\frac{1}{|\mathbf{Y}|}\sum_{\mathbf{y}_i\in{\mathbf{Y}}}{\rho}_{\mu}\left(d_{\mathbf{y}_i,\tilde{\mathbf{X}}\left(\mathbf{w}\right)}\right)\bigg],
	\end{aligned}
\end{equation}
where $\tilde{\mathbf{X}}\left(\mathbf{w}\right)$ denotes the transformed source point cloud after performing the transformation predicted by our deep model on the source point cloud $\mathbf{X}$. 
Benefitting from the 

\begin{table*}[t]
	\centering
	\resizebox{1.0\linewidth}{!}{
		\begin{tabular}{lcccc|cccc|cccc}
			\toprule[1.5pt]
			\multirow{2}{*}{\textit{Model}} & \multicolumn{4}{c|}{\textit{Unseen object}} & \multicolumn{4}{c|}{\textit{Unseen category}} & \multicolumn{4}{c}{\textit{Unseen object with noise}} \\
			& {RMSE(R)}& {RMSE(t)}& {MAE(R)} & {MAE(t)} & {RMSE(R)}& {RMSE(t)}& {MAE(R)} & {MAE(t)} &  {RMSE(R)}& {RMSE(t)}& {MAE(R)} & {MAE(t)}            \\ \midrule 
			ICP~\cite{besl1992method} ($\ast$) & 13.7952 &  0.0391 & 4.4483 & 0.0196& 14.7732&0.0351&3.5938&0.0132& 12.5413 & 0.0398 & 4.2826 & 0.0184 \\ 
			Go-ICP~\cite{yang2013go} ($\ast$) & 14.7223 & 0.0328 & 3.5112 & 0.0127 &13.8322& 0.0321&3.1579&0.0121&14.5225 & 0.0329 & 3.4252 & 0.0114  \\ 
			Super4PCS~\cite{mellado2014super} ($\ast$) & \underline{1.5764} & \underline{0.0034} & {5.3512} & 0.0214 & 2.5186 & \underline{0.0032} &1.4364 & 0.0041 &12.4551 & 0.0201 & 1.3780 &  0.0033 \\ 
			SDRSAC ~\cite{le2019sdrsac} ($\ast$) & 3.9173& 0.0121 & 2.7956 & 0.0102& 4.2475 & 0.0139&3.0144&0.0121&\underline{0.4719} & \underline{0.0185} & 2.3965 & 0.0164 \\
			FGR~\cite{zhou2016fast} ($\ast$) & 3.7055 & 0.0088 & 0.5972 & \underline{0.0020}&3.1251&0.0074&\underline{0.4469} &\underline{0.0013} & \textbf{0.4712} & 0.0703 & \textbf{0.2353} & 0.0404  \\ 
			DCP-v2~\cite{wang2019deep} ($\lozenge$) & 4.8962 & 0.0248 & 3.3297 & 0.0169 &6.3787&0.0246&4.4222&0.0173&5.1575 & 0.0251 & 3.4708 & 0.0176 \\ 
			IDAM~\cite{li2020iterative} ($\lozenge$) & 2.3384 & 0.0102 & \underline{0.4711} & 0.0025&\underline{2.1566} &0.0151&0.6135&0.0037&3.5701 & 0.0206 & 1.0642 & \underline{0.0066} \\
			FMR~\cite{huang2020feature} ($\triangle$) &9.0997 & 0.0204 & 3.6497 & 0.0101&9.1322& 0.0223&3.8593&0.0113&5.5605 & {0.0194} & 2.5437 & 0.0072 \\ 
			\midrule
			CEMNet (ours, $\triangle$) & \textbf{1.5018} & \textbf{0.0009} & \textbf{0.1385} & \textbf{0.0001} & \textbf{1.1013} & \textbf{0.0020} & \textbf{0.0804} & \textbf{0.0002} & {2.2722} & \textbf{0.0014} & \underline{0.3799} & \textbf{0.0008} \\ \bottomrule[1.5pt]
	\end{tabular}}
	\vspace{0.5mm}
	\caption{Comparison results on partially overlapping \textit{ModelNet40} dataset. $(\ast)$, $(\triangle)$ and $(\lozenge)$ denote the traditional, unsupervised and fully supervised deep methods, respectively.}\label{tab:modelnet40}
	\vspace{-2mm}
\end{table*}

\section{Experiment}
In this section, we perform extensive experiments and ablation studies on benchmark datasets, including the \textit{ModelNet40} \cite{wu20153d}, \textit{7Scene} \cite{shotton2013scene} and \textit{ICL-NUIM}~\cite{handa2014benchmark}.  
We simply name our \textbf{CEM} based {R}egistration \textbf{Net}work as \textbf{CEMNet}. 


\subsection{Implementation Details}
\label{sec:detail}
We train our model using Adam optimizer with learning rate $10^{-4}$, and weight decay $5\times 10^{-4}$ for $50$ epochs. The batch size is set to $32$.
In our differentiable CEM module,  we set the numbers of iterations $T$ and candidates $N$ to $15$ and $1000$, respectively, and the maximum allowable distance $\varepsilon$ is set to $0.1$. 
For the fused reward for transformation evaluation, we set $M$ to $3$ and the impact weight $\alpha$ in the score function is set to $0.5$. 
The hyper-parameter $\mu$ in the Geman-McClure estimator used in our loss function is set to $0.01$.
We utilize PyTorch to implement our project and perform all experiments on the server equipped with a 2080Ti GPU and an Intel i5 2.2GHz CPU.

\vspace{-1mm}
\subsection{Evaluation on ModelNet40}
\vspace{-1mm}
\label{sec:modelnet40}
We first test our method on the \textit{ModelNet40} datatset~\cite{wu20153d}, which contains $40$ categories and is constructed via uniformly sampling $1,024$ points from each of $12,311$ CAD models. 
Then, we divide it into two parts where $2,468$ models are used for testing and the remaining models are for training.
Following the setting in \cite{wang2019deep}, we first generate a random rotation matrix and translation vector via uniformly sampling in the Euler angle range $\left[0,45^{\circ}\right]$ and the translation range $\left[-0.5, 0.5\right]$. Then, we use the transformed source point cloud as the target point cloud and the generated transformation is as the ground truth. 
The metric for evaluating the registration precision contains the \textit{root mean squared error (RMSE)} and \textit{mean absolute error (MAE)} between the predicted transformation and the ground-truth transformation, where the error about Euler angle uses the degree as the unit. We compare our method with eight state-of-the-art methods, where ICP~\cite{besl1992method}, Go-ICP~\cite{yang2013go}, Super4PCS~\cite{mellado2014super}, SDRSAC \cite{le2019sdrsac} and FGR~\cite{zhou2016fast} belong to the traditional methods, and IDAM~\cite{li2020iterative} and DCP-v2~\cite{wang2019deep} are fully supervised deep methods while FMR~\cite{huang2020feature} is unsupervised. Some visualization results are presented in Fig.~\ref{fig:vis} and more qualitative examples can be seen in Appendix D.

\noindent\textbf{Partially overlapping unseen object.} 
We first evaluate our method on partially overlapping unseen objects, where the test and training datasets contain all $40$ categories, and the non-overlapping region may still exist in the perfectly aligned point clouds.  
Following \cite{wang2019prnet}, to construct the partially overlapping point clouds, we first randomly sample a point in source and target point clouds, respectively. Then, we perform the farthest point subsampling (FPS) to sample $75\%$ ($768$) points while the remaining $25\%$ points are viewed as the missing points and discarded. 
We utilize the implementations provided by the authors to train FMR, DCP-v2, and IDAM on our modified training dataset with partial overlap.
The left column of Table \ref{tab:modelnet40} shows that compared to other algorithms, our unsupervised method can obtain superior registration performance with all evaluation metrics and even exceeds the fully supervised DCP-v2 and IDAM, which owes to that the CEM module guided by the sampling network can effectively search for the accurate transformation.

\begin{table}[htbp]
	\centering
	\resizebox{1.0\linewidth}{!}{
		\begin{tabular}{llcccc}
			\toprule[1.5pt]
			Dataset & Model & RMSE(R) & RMSE(t) & MAE(R) & MAE(t) \\ \midrule 
			\multirow{9}{*}{\textit{7Scene}} & ICP~\cite{besl1992method} ($\ast$) & 19.9166 &  0.1127 & 7.5760 & 0.0310 \\ 
			&Go-ICP~\cite{yang2013go} ($\ast$) & 24.2743 & 0.0360 & 7.1068 & 0.0137  \\ 
			&Super4PCS~\cite{mellado2014super} ($\ast$) & 19.2603 & 0.3646 & 15.7001 & 0.2898 \\ 
			&SDRSAC ~\cite{le2019sdrsac} ($\ast$) & 0.3501 & 0.4997 & 0.2925 & 0.4997 \\
			&FGR~\cite{zhou2016fast} ($\ast$) & \underline{0.2724} & \textbf{0.0011} & \underline{0.1380} & \underline{0.0006} \\ 
			&DCP-v2 ~\cite{wang2019deep} ($\lozenge$) & 7.5548 & 0.0411 & {5.6991} & {0.0303} \\
			&IDAM~\cite{li2020iterative} ($\lozenge$) & 10.5306 & 0.0539 & 5.6727 & 0.0303 \\ 
			&FMR~\cite{huang2020feature} ($\triangle$) & 8.6999 & 0.0199 & 3.6569 & 0.0101 \\ \cmidrule{2-6}
			&CEMNet (ours, $\triangle$) & \textbf{0.1768} & \underline{0.0012} & \textbf{0.0434} & \textbf{0.0002}     \\ \midrule
			\multirow{9}{*}{\textit{ICL-NUIM}} & ICP~\cite{besl1992method} ($\ast$) & 10.1247 & 0.3006 & 2.1484 & 0.0693 \\
			&Go-ICP~\cite{yang2013go} ($\ast$)  & \underline{1.5514} & 0.0601 & \underline{0.6333} & 0.0241 \\ 
			&Super4PCS~\cite{mellado2014super} ($\ast$) & 28.8616 & 0.3091 & 24.1373 & 0.2502 \\ 
			&SDRSAC ~\cite{le2019sdrsac} ($\ast$) & 9.4074 & 0.2477 & 7.8627 & 0.2076 \\
			&FGR~\cite{zhou2016fast} ($\ast$)  & {3.0423} & {0.1275} & {1.9571} & {0.0659} \\
			&DCP-v2 ~\cite{wang2019deep} ($\lozenge$) & 9.2142 & \underline{0.0191} & {6.5826} & \underline{0.0134} \\
			&IDAM~\cite{li2020iterative} ($\lozenge$) & 9.4539 & 0.3040 & 4.4153 & 0.1385 \\
			&FMR~\cite{huang2020feature} ($\triangle$) & 1.8282 & 0.0685 & 1.1085 & 0.0398 \\  
			\cmidrule{2-6}
			&CEMNet (ours, $\triangle$) &  \textbf{0.0821} & \textbf{0.0002} & \textbf{0.0211} & \textbf{0.0001}\\ \bottomrule[1.5pt]
	\end{tabular}}
	\vspace{0.5mm}
	\caption{Comparison results on \textit{7Scene} and \textit{ICL-NUIM} datasets. $(\ast)$, $(\triangle)$ and $(\lozenge)$ denote the traditional, unsupervised and fully supervised deep methods, respectively.}	\label{table:indoor}
	\vspace{-6mm}
\end{table}

\noindent\textbf{Partially overlapping unseen category.} To further test the generalization ability of our model, we utilize $20$ categories for training and then test the trained model on another $20$ unseen categories.
The comparison results are listed in the middle column of Table~\ref{tab:modelnet40}. Our method presents good generalization ability on unseen categories and still obtains the best scores on all criteria. 

\noindent\textbf{Partially overlapping unseen object with noise.} For the robustness evaluation of our method in the presence of noise, we jitter each point of the partially overlapping source and target point clouds with Gaussian noise. Following \cite{wang2019prnet}, we sample the noise in each axis, clipped by $\left[-0.05, 0.05\right]$, from a Gaussian distribution with the mean $0$ and the standard deviation $0.01$.  
The right column of Table~\ref{tab:modelnet40} demonstrates that although it is worse than FGR in terms of the \textit{RMSE} and \textit{MAE} criteria with respect to rotation, it yields better performance on translation.

\subsection{Evaluation on 7Scene and ICL-NUIM}
We further evaluate our method on two indoor datasets: \textit{7Scene} and \textit{ICL-NUIM}. The former consists of seven scenes: Chess, Fires, Heads, Office, Pumpkin, Redkitchen, and Stairs, and is split into two parts, one containing $296$ scans for training and the other containing $57$ scans as the test dataset.
The latter is split into $1,278$ and $200$ scans for training and test, respectively. 
As in Sec.~\ref{sec:modelnet40}, we exploit the FPS operation to obtain the partial point clouds and use the transformed source point cloud as the target point cloud, where the transformation is also generated via randomly sampling. As demonstrated in Table~\ref{table:indoor}, on \textit{ICL-NUIM}, our method exhibits higher registration precision on all criteria. In addition, on \textit{7Scene}, our method is slightly worse than FGR in terms of the \textit{RMSE} of translation but shows the lower alignment errors on other criteria.

\begin{figure}[t]
	\centering  
	\subfigure[\# of candidates and iterations]{
		\label{summary_3}
		\includegraphics[width=0.485\columnwidth]{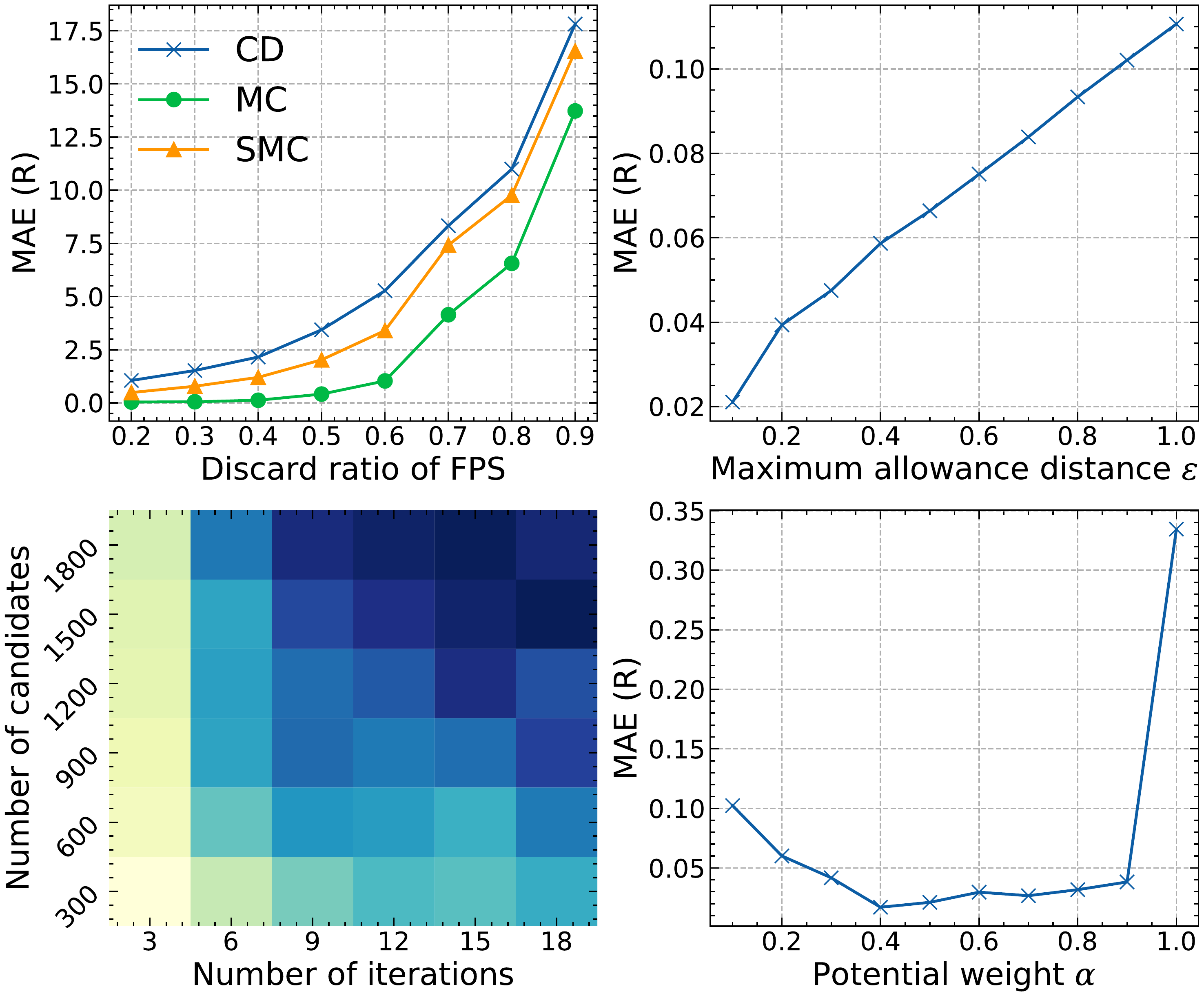}}
	\subfigure[ $\alpha$ in the fused score function]{
		\label{summary_4}
		\includegraphics[width=0.478\columnwidth]{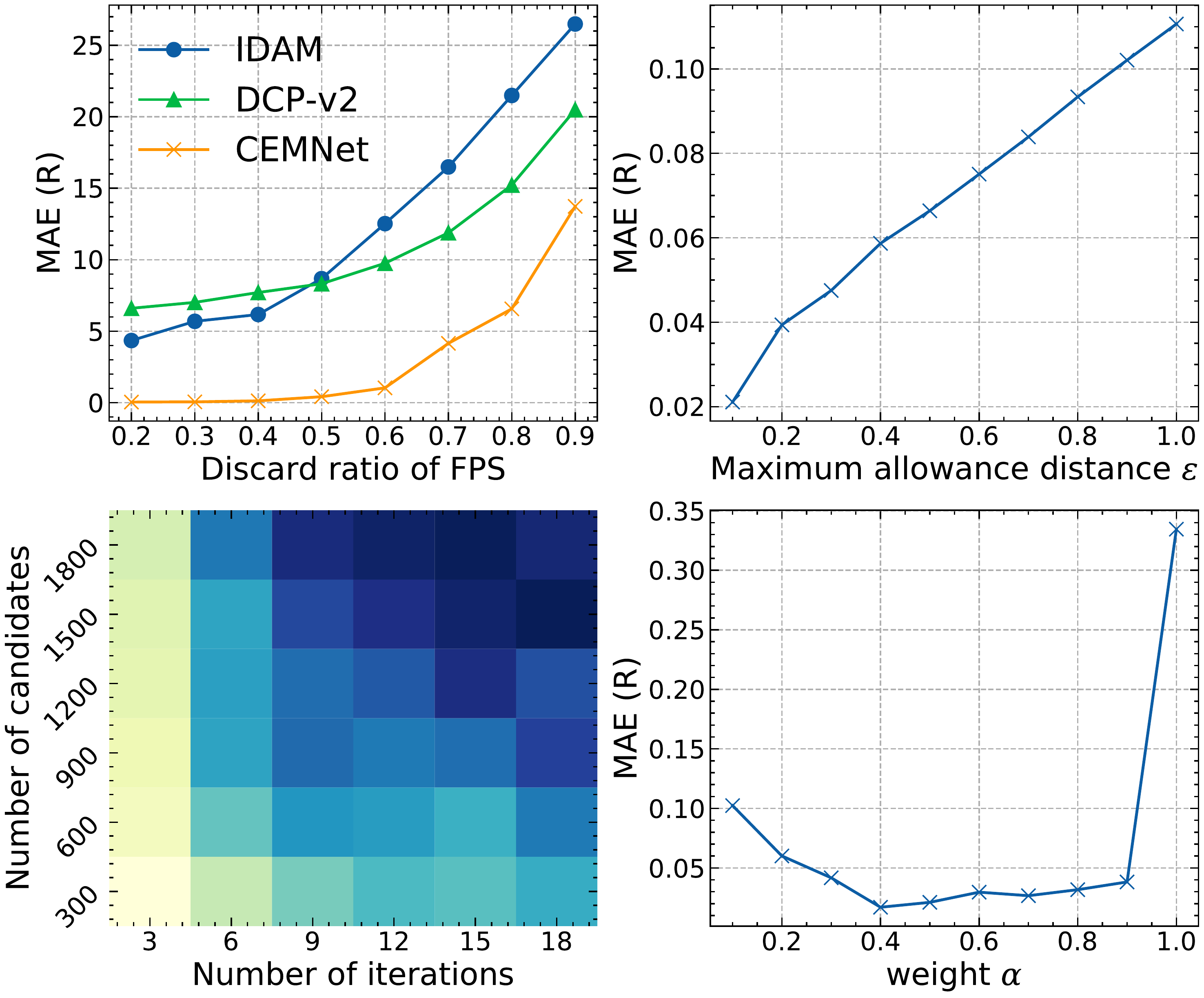}}
	\caption{Rotation errors in the cases of (a) different numbers of candidates and iterations, and (b) different weights in fused score function on \textit{ICL-NUIM} dataset.}
	\label{summary}
	\vspace{-2mm}
\end{figure}

\subsection{Ablation Study and Analysis}
\vspace{-2mm}
\label{sec:abl}
\begin{table}[]
	\centering
	\resizebox{1\columnwidth}{!}{
		\begin{tabular}{ccccccc}
			\toprule[1.5pt]
			&&&\multicolumn{2}{c}{\textit{{ModelNet40}$^{uc}$}} & \multicolumn{2}{c}{\textit{ICL-NUIM}}\\\cmidrule{4-7}
			\textit{SN} & \textit{Future} & \textit{dCEM}
			&MAE(R)& MAE(t)&MAE(R)&MAE(t) \\ \midrule 
			$\surd$ & & & {4.4222} & {0.0173} & {6.5826} & {0.0134} \\ 
			& &$\surd$ & {1.8625} & {0.0041} & {1.8333} & {0.0023} \\ 
			&$\surd$ &$\surd$ & {0.3675} & \underline{0.0004} & {0.8364} & {0.0048} \\ 
			$\surd$& &$\surd$ & \underline{0.1781} & \underline{0.0004} & \underline{0.3345} & \underline{0.0005} \\ 
			$\surd$ &$\surd$ &$\surd$ & \textbf{0.0804} & \textbf{0.0002} & \textbf{0.0211} & \textbf{0.0001} \\ 
			\bottomrule[1.5pt]
	\end{tabular}}
	\vspace{0.5mm}
	\caption{Ablation study of different components on \textit{ModelNet40$^{uc}$} (unseen category) and \textit{ICL-NUIM} datasets. \textit{SN}: Sampling network module; \textit{Future}: ICP driven future reward evaluation; \textit{dCEM}: differentiable CEM module.}
	\label{different2}
	\vspace{-5mm}
\end{table}


\noindent\textbf{Sampling network module.} As shown in the fourth and fifth rows of Table \ref{different2}, equipped with the sampling network module (\textit{SN}), \textit{dCEM} (differentiable CEM) and \textit{dCEM}+\textit{Future} (differentiable CEM with future reward) can obtain significant error reduction on all criteria. This is because that \textit{SN} can learn a better initial sampling distribution, which assists the \textit{dCEM} to efficiently search for the optimal solution. Note that we use the Gaussian distribution $\mathcal{N}\left(\boldsymbol{0},\boldsymbol{1}\mathbb{I}\right)$ as the pre-defined sampling distribution for our method without \textit{SN}. 
Moreover, since the sampling network has a similar model design as DCP-v2 except for the branch used for variance prediction, the results in the first row refer to the scores of the fully supervised DCP-v2. One can see that without planning in the CEM module, \textit{SN} yields poor performance, even guided by the ground truth transformation, which implies the necessity of the proposed CEM module for the precise registration.  

\noindent\textbf{Fused score function.} We further evaluate the effect of ICP based future reward in the fused score function of CEM. As shown in the second and fourth rows of Table~\ref{different2}, the performance drops significantly without considering the future reward of the transformation candidate. Furthermore, Table~\ref{different3} also shows that as the iteration times $M$ of using the future reward increases, the error keeps decreasing while the time cost continues to increase. To balance the performance and speed, we set $M$ to $3$ in all experiments. Finally, we test the fluctuation of the error with different weights $\alpha$ in the fused function. Fig.~\ref{summary_4}  shows that a good balance between the current and future rewards tends to bring higher precision. 

\begin{table}[]
	\centering
	\resizebox{1\columnwidth}{!}{
		\begin{tabular}{cccccc}
			\toprule[1.5pt]
			&&\multicolumn{2}{c}{\textit{{ModelNet40}$^{uc}$}} & \multicolumn{2}{c}{\textit{{ModelNet40}$^{uo}$}}\\\cmidrule{3-6}
			\textit{\# Future (M)} & \textit{Time (ms)} & MAR(R) & MAR(t)
			&MAE(R)& MAE(t) \\ \midrule 
			0 & \textbf{184.7} & 0.1782 & {0.00040} & {0.3505} & {0.00053} \\ 
			1 & \underline{235.5} & 0.1709 & {0.00021} & {0.2688} & {0.00018} \\ 
			2 & 263.7 & {0.1198} & {0.00019} & {0.2137} & {0.00015} \\ 
			3 & 309.5 & \underline{0.0804} & \underline{0.00015} & \underline{0.1385} & \underline{0.00012}  \\ 
			4 & 335.2 & \textbf{0.0470} & \textbf{0.00011} & \textbf{0.1258} & \textbf{0.00010}  \\ 
			\bottomrule[1.5pt]
	\end{tabular}}
	\vspace{0.5mm}
	\caption{The time and precision vary with different number of iterations using ICP driven potential evaluation on  \textit{ModelNet40$^{uc}$} (unseen category) and \textit{ModelNet40$^{uo}$} (unseen object) datasets.}
	\label{different3}
	\vspace{-5mm}
\end{table}


\noindent\textbf{Parameters in CEM.}
We test the sensitivity of the hyper-parameters in CEM, including the numbers of the iterations $T$ and the candidates $N$.
Fig.~\ref{summary_3} shows that as $T$ and $N$ increase, the error tends to decrease (from the light color to the dark color). However, as shown in Appendix C, their performance gain is at the cost of the inference time and we set $N$ and $T$ to $1000$ and $10$ in all experiments.

\noindent\textbf{Inference time.} 
We use the averaged running time on each case from \textit{ModelNet40} dataset (unseen object) to evaluate the inference time of each algorithm. Note that the traditional registration methods are performed on Intel i5 2.2GHz CPU while the deep learning based methods are executed on 2080Ti GPU. 
The time cost of our method is $310$ms, while the cost of the other methods are ICP ($14$ms), Go-ICP ($2,278$ms), Super4PCS ($6,790$ms), FGR ($75$ms), SDRSAC ($23,651$ms), FMR ($414$ms), DCP-v2 ($34$ms), and IDAM ($53$ms), respectively.

\vspace{-3mm}
\section{Conclusion}
\vspace{-2mm}
In this paper, we proposed a novel sampling network guided cross-entropy method for unsupervised point cloud registration.
In our framework, the sampling network is used to provide a prior sampling distribution for CEM. Guided by this learnable distribution, the CEM module further heuristically searches for the optimal transformation. During this optimization process, we designed a fused score function combining current and ICP based future rewards for more accurate transformation evaluation. Furthermore, we also replaced the hard top-$k$ selection with the differentiable sparsemax function for the end-to-end model training.

%

\vspace{-2mm}
\section*{Acknowledgments}
\vspace{-1mm}
This work was supported by the National Science Fund of China (Grant Nos. U1713208, 61876084, 61876083, 62072242).

{\small
\bibliographystyle{ieee_fullname}
\bibliography{paper}
}

\end{document}